\documentclass[letterpaper]{article} 
\usepackage{aaai24}  
\usepackage{times}  
\usepackage{helvet}  
\usepackage{courier}  
\usepackage[hyphens]{url}  
\usepackage{graphicx} 
\usepackage{natbib}  
\usepackage{caption} 
\usepackage{algorithm}
\usepackage{algorithmic}

\usepackage{cuted}

\usepackage{tcolorbox}

\usepackage{amsmath}

\usepackage{times}
\usepackage{epsfig}
\usepackage{graphicx}
\usepackage{amsmath}
\usepackage{amssymb}

\usepackage{lipsum} 

\usepackage{graphicx}
\usepackage{amsmath}
\usepackage{amssymb}
\usepackage{booktabs}

\usepackage{tikz}
\newcommand*{\circled}[1]{\lower.7ex\hbox{\tikz\draw (0pt, 0pt)%
    circle (.45em) node {\makebox[1em][c]{\small #1}};}}

\usepackage{xr}

\usepackage{array}
\usepackage{mathrsfs}
\usepackage{amsmath}
\usepackage{bm}
\usepackage[cmintegrals]{newtxmath}
\usepackage{xcolor}

\usepackage{epsfig}
\usepackage{graphicx}
\usepackage{amsmath}
\usepackage{amssymb}
\DeclareMathOperator*{\argmax}{arg\,max}

\usepackage{dsfont}
\usepackage{amsfonts}
\usepackage[switch]{lineno}
\usepackage{enumitem}

\usepackage{float}

\usepackage{bbding}
\usepackage{booktabs}
\usepackage{multirow}
\usepackage{color}

\usepackage{subcaption}
\usepackage{xcolor}

\usepackage{newfloat}
\usepackage{listings}
\DeclareCaptionStyle{ruled}{labelfont=normalfont,labelsep=colon,strut=off} 
\floatstyle{ruled}
\newfloat{listing}{tb}{lst}{}
\floatname{listing}{Listing}
\title{FedDiv: Collaborative Noise Filtering for Federated Learning with Noisy Labels}

\author{
    Jichang Li\textsuperscript{\rm 1,\rm 2},
    Guanbin Li\textsuperscript{\rm 1,\rm 3}{\Large \thanks{Corresponding Authors are Guanbin Li and Yizhou Yu.}},
    Hui Cheng\textsuperscript{\rm 4},
    Zicheng Liao\textsuperscript{\rm 2,\rm 5},
    Yizhou Yu\textsuperscript{\rm 2{\small *}}
}
\affiliations{
    \textsuperscript{\rm 1}School of Computer Science and Engineering, Research Institute of Sun Yat-sen University in Shenzhen, Sun Yat-sen University, Guangzhou, China\\
    \textsuperscript{\rm 2}Department of Computer Science, The University of Hong Kong, Hong Kong\\
    \textsuperscript{\rm 3}Guangdong Province Key Laboratory of Information Security Technology\\
    \textsuperscript{\rm 4}UM-SJTU Joint Institute, Shanghai Jiao Tong University, Shanghai, China\\
    \textsuperscript{\rm 5}Zhejiang University, Hangzhou, China\\

    {csjcli@connect.hku.hk, liguanbin@mail.sysu.edu.cn, chenghui\_123@sjtu.edu.cn\\  zichengliao@gmail.com, yizhouy@acm.org}
    
}

\usepackage{bibentry}

\begin{document}


\maketitle

\begin{abstract}
Federated Learning with Noisy Labels (F-LNL) aims at seeking an optimal server model via collaborative distributed learning by aggregating multiple client models trained with local noisy or clean samples. On the basis of a federated learning framework, recent advances primarily adopt label noise filtering to separate clean samples from noisy ones on each client, thereby mitigating the negative impact of label noise. However, these prior methods do not learn noise filters by exploiting knowledge across all clients, leading to sub-optimal and inferior noise filtering performance and thus damaging training stability. In this paper, we present \texttt{FedDiv} to tackle the challenges of F-LNL. Specifically, we propose a global noise filter called Federated Noise Filter for effectively identifying samples with noisy labels on every client, thereby raising stability during local training sessions. Without sacrificing data privacy, this is achieved by modeling the global distribution of label noise across all clients. Then, in an effort to make the global model achieve higher performance, we introduce a Predictive Consistency based Sampler to identify more credible local data for local model training, thus preventing noise memorization and further boosting the training stability. Extensive experiments on CIFAR-10, CIFAR-100, and Clothing1M demonstrate that \texttt{FedDiv} achieves superior performance over state-of-the-art F-LNL methods under different label noise settings for both IID and non-IID data partitions. Source code is publicly available at https://github.com/lijichang/FLNL-FedDiv.
\end{abstract}

\begin{figure}[ht]
    \centering
    \includegraphics[width=0.45\textwidth]{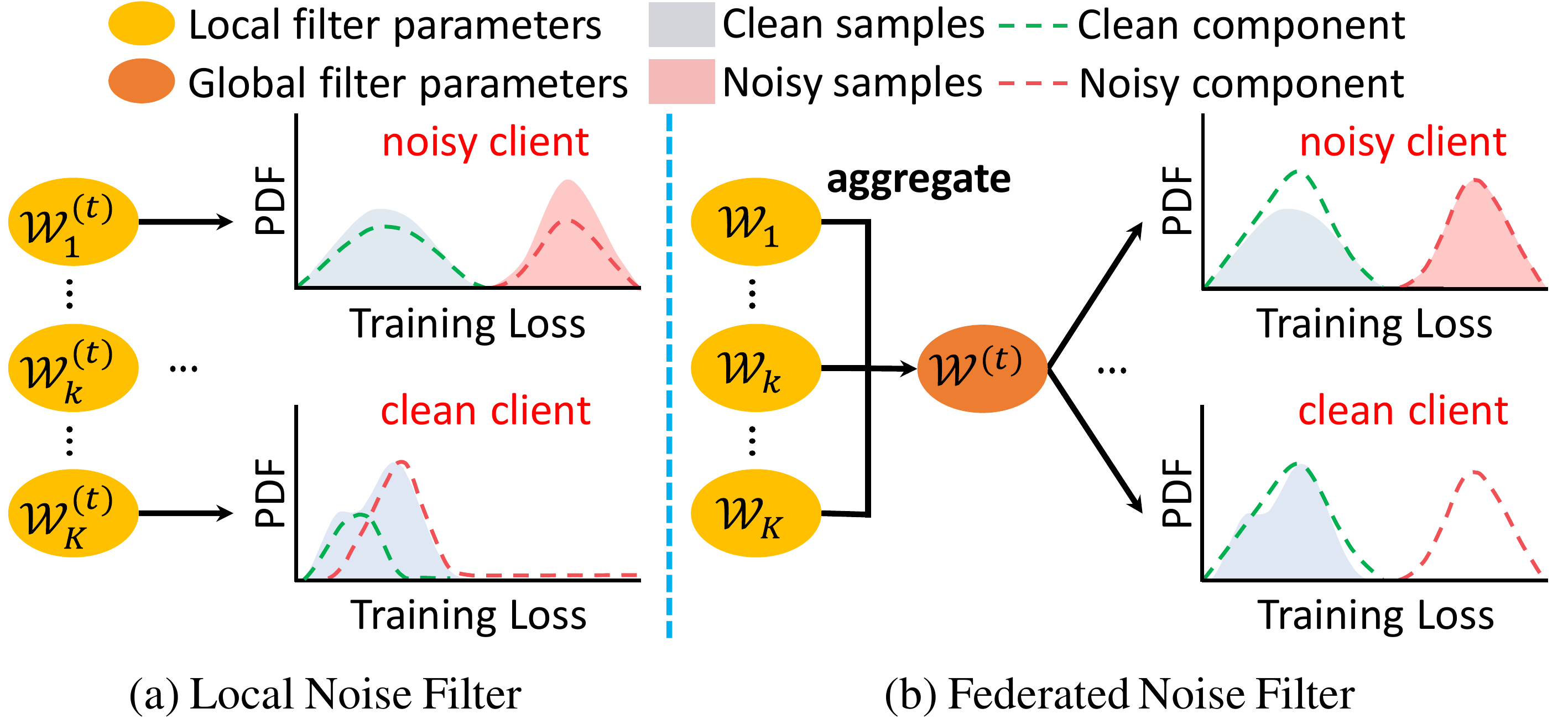}
    \caption{
    (a) Local noise filtering may have limited capabilities as each client develops its own local noise filter using its own private data only. Especially on clean clients, such filters would incorrectly identify a subset of clean samples to be noisy.
    (b) Collaborative noise filtering proposed by us significantly improves the performance of label noise filtering on each client as a federated noise filter is learned by distilling knowledge from all clients.
    PDF: Probability density function.} 
    \label{Figure:Overview}
\end{figure}

\section{Introduction}
 
Compared to traditional Centralized Learning~\cite{li2019relation, li2021cross, li2019semi, wu2019mutual, wu2019enhancing, huang2023divide}, Federated Learning (FL) is a novel paradigm facilitating collaborative learning across multiple clients without requiring centralized local data~\cite{FedAvg}. Recently, FL has shown significant real-world success in areas like healthcare~\cite{healthcare1}, recommender systems~\cite{recommendation1}, and smart cities~\cite{smartcity1}. However, these FL methods presume clean labels for all client's private data, which is often not the case in reality due to data complexity and uncontrolled label annotation quality~\cite{tanno2019learning, kuznetsova2020open}. Especially with the blessing of privacy protection, it is impossible to ensure absolute label accuracy. Therefore, this work centers on Federated Learning with Noisy Labels (F-LNL). In F-LNL, a global neural network model is fine-tuned via distributed learning across multiple local clients with noisy samples. Like~\cite{xu2022fedcorr},  we here also assume some local clients have noisy labels (namely noisy clients), while others have only clean labels (namely clean clients).
 
Besides the private data on local clients, F-LNL faces two primary challenges: data heterogeneity and noise heterogeneity~\cite{kim2022fedrn,RoFL}. Data heterogeneity refers to the statistically heterogeneous data distributions across different clients, while noise heterogeneity represents the varying noise distributions among clients. It has been demonstrated by~\cite{xu2022fedcorr,kim2022fedrn} that these two challenges in F-LNL may lead to instability during local training sessions. 
Previous studies in F-LNL~\cite{xu2022fedcorr,kim2022fedrn} have demonstrated that many FL approaches, such as~\cite{FedAvg, FedProx}, are now in use to adequately address the first challenge. These approaches primarily focus on achieving training stability with convergence guarantees by aligning the optimization objectives of local updates and global aggregation. However, they do not address label noise on individual clients. Therefore, to tackle noise heterogeneity, such F-LNL algorithms propose separating noisy data into clean and noisy samples, occasionally complemented by relabeling the noisy samples. This aims to mitigate the negative effects of noisy labels and prevent local model overfitting to such label noise, avoiding severe destabilization of the local training process.

Some F-LNL methods~\cite{kim2022fedrn,xu2022fedcorr} thus emphasize proposing local noise filtering, where each client develops its own noise filter to identify noisy labels. However, these noise filtering strategies omit the potential of learning prosperous knowledge from other clients to strengthen their capacities. Instead, they rely heavily on each client’s own private data for training, thus leading to sub-optimal and inferior performance. For example, as shown in Figure~\ref{Figure:Overview}(a), limited training samples available on each client impede the accurate modeling of their local noise distributions, significantly restricting noise filtering capabilities. In addition, if noise filtering and relabeling are not handled properly, overfitting of noisy labels can inevitably occur, leading to noise memorization and thus degrading global performance upon model aggregation. Developing strategies to prevent noise memorization~\cite{noisememory6} while enhancing training stability is critical, yet existing F-LNL algorithms have not succeeded in achieving this.
 
In this paper, we present a novel framework, \texttt{FedDiv}, to address the challenges of F-LNL. To perform label noise filtering per client, \texttt{FedDiv} consists of a global noise filter, called Federated Noise Filter (FNF), constructed by modeling the global distribution of noisy samples across all clients. Specifically, by fitting the loss function values of private data, the parameters of a local Gaussian Mixture Model (GMM) can be iteratively learned on each client to model its local noise distributions. These parameters are then aggregated on the server to construct a global GMM model that serves as a global noise filter, effectively classifying samples on each local client into clean or noisy. As depicted in Figure~\ref{Figure:Overview}(b), by leveraging collaboratively learned knowledge across all clients, \texttt{FedDiv} demonstrates a robust capability to fit the local label noise distributions within individual clean or noisy clients. This ability enhances noise filtering performance per client, and thus reduces training instability during local training sessions, while preserving data privacy.

\begin{figure*}[t]
    \centering
    \includegraphics[width=0.95\textwidth]{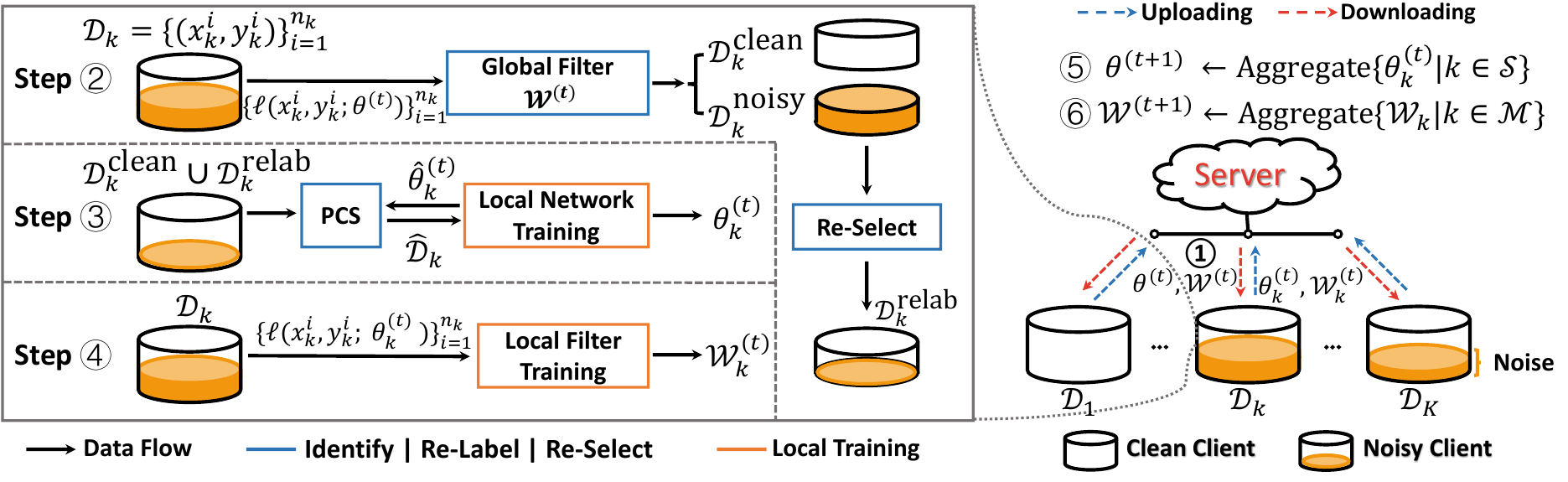}
    \caption{An overview of the training procedure proposed by FedDiv.
    In this work, the parameters of a local neural model and a local noise filter are simultaneously learned on each client during the local training sessions, while both types of parameters are aggregated on the server.}
    \label{Figure:Framework}
\end{figure*}

After label noise filtering, we remove noisy labels from the identified noisy samples on each client and relabel those samples exhibiting high prediction confidence using the pseudo-labels predicted by the global model. To further prevent local models from memorizing label noise and improve training stability, we introduce a Predictive Consistency based Sampler (PCS) for identifying credible local data for local model training. Specifically, we enforce the consistency of class labels predicted by both global and local models, and apply counterfactual reasoning~\cite{holland1986statistics, debiasedlearning} to generate more reliable predictions for local samples.

In summary, the contributions of this paper are as follows.
\begin{itemize}
    \item We propose a novel one-stage framework, \texttt{FedDiv}, for addressing the task of Federated Learning with Noisy Labels (F-LNL). To enable stable training, \texttt{FedDiv} learns a global noise filter by distilling the complementary knowledge from all clients while performing label noise filtering locally on every client.
    \item We introduce a Predictive Consistency based Sampler to perform labeled sample re-selection on every client, thereby preventing local models from memorizing label noise and further improving training stability.
    \item Through extensive experiments conducted on CIFAR-10~\cite{cifar}, CIFAR-100~\cite{cifar}, and Clothing1M~\cite{clothing1m} datasets, we demonstrate that \texttt{FedDiv} significantly outperforms state-of-the-art F-LNL methods under various label noise settings for both IID and non-IID data partitions.
\end{itemize}

\section{Related Work}

\noindent
\textbf{Centralized Learning with Noisy Labels (C-LNL).}
Diverging from conventional paradigms for centralized learning, e.g.~\cite{li2023betweenness, li2023idm}, which operates on training samples only with clean labels, several studies have demonstrated the effect of methods to address C-LNL in reducing model overfitting to noisy labels. For instance, \texttt{JointOpt}~\cite{JointOpt} proposed a joint optimization framework to correct labels of noisy samples during training by alternating between updating model parameters and labels. As well, \texttt{DivideMix}~\cite{li2019dividemix} dynamically segregated training examples with label noise into clean and noisy subsets, incorporating auxiliary semi-supervised learning algorithms for further model training. Other strategies for handling C-LNL tasks include estimating the noise transition matrix~\cite{cheng2022instance}, reweighing the training data~\cite{ren2018learning}, designing robust loss functions~\cite{englesson2021generalized}, ensembling existing techniques~\cite{li2022neighborhood}, and so on.

Given privacy constraints in decentralized applications, the server cannot directly access local samples of all clients to construct centralized noise filtering algorithms. Besides, a limited number of private data on local clients may also restrict the noise filtering capability. Hence, despite the success of existing C-LNL algorithms, they may no longer be feasible in federated settings~\cite{xu2022fedcorr}. 

\noindent
\textbf{Federated Learning with Noisy Labels.}
Numerous methods address challenges in federated scenarios with label noise. For instance, \texttt{FedRN}~\cite{kim2022fedrn} detects clean samples in local clients using ensembled Gaussian Mixture Models (GMMs) trained to fit loss function values of local data assessed by multiple reliable neighboring client models. \texttt{RoFL} \cite{RoFL} directly optimizes using small-loss instances of clients during local training to mitigate label noise effects. Meanwhile, \texttt{FedCorr}\cite{xu2022fedcorr} first introduces a dimensionality-based noise filter to segregate clients into clean and noisy groups using local intrinsic dimensionalities~\cite{LID}, and trains local noise filters to separate clean examples from identified noisy clients based on per-sample training losses.

However, existing F-LNL algorithms concentrate on local noise filtering, utilizing each client's private data but failing to exploit collective knowledge across clients. This limitation might compromise noise filtering efficacy, resulting in incomplete label noise removal and impacting the stability of local training sessions. Conversely, \texttt{FedDiv} proposes distilling knowledge from all clients for federated noise filtering, enhancing label noise identification in each client's sample and improving FL model training amidst label noise.

\section{Methodology}

In this section, we introduce the proposed one-stage framework named \texttt{FedDiv} for federated learning with noisy labels. In detail, we first adopt the classic FL paradigm, namely \texttt{FedAvg}~\cite{FedAvg}, to train a neural network model. On the basis of this FL framework, we propose a global filter model called Federated Noise Filter (FNF) to perform label noise filtering and noisy sample relabeling on every client. Then, to improve the stability of local training, a Predictive Consistency based Sampler (PCS) is presented to conduct labeled sample re-selection, keeping client models from memorizing label noise. 
 
Let us consider an FL scenario with one server and $K$ local clients denoted by $\mathcal{M}$. Each client $k\in\mathcal{M}$ has its own private data consisting of $n_k$ sample-label pairs $\mathcal{D}_{k}=\{(x_{k}^i, y_{k}^i)\}_{i=1}^{n_k}$, where $x_{k}^i$ is a training sample, and $y_{k}^i$ is a label index over $C$ classes. Here, we divide local clients into two groups: clean clients (with noise level $\delta_k=0$ where $k\in\mathcal{M}$), which only have samples with clean labels, and noisy clients (with $\delta_k>0$), whose private data might have label noise at various levels. Also, in this work, both IID and non-IID heterogeneous data partitions are considered. 

\begin{table*}[t]
    \centering
    \small
    \begin{tabular}{l|ccrrrc}
        \toprule
        \midrule
        \multicolumn{1}{c|}{\textbf{Method}} & \multicolumn{6}{c}{\textbf{Best Test Accuracy $\pm$ Standard Deviation}} \\
        & \multicolumn{2}{c}{$\rho$=0.4} & \multicolumn{2}{c}{$\rho$=0.6} & \multicolumn{2}{c}{$\rho$=0.8} \\
        & $\tau$=0.0 & $\tau$=0.5 & $\tau$=0.0 & $\tau$=0.5 & $\tau$=0.0 & $\tau$=0.5 \\
        \midrule
        FedAvg       & 89.46$\pm$0.39 & 88.31$\pm$0.80 & 86.09$\pm$0.50 & 81.22$\pm$1.72 & 82.91$\pm$1.35 & 72.00$\pm$2.76 \\
        RoFL         & 88.25$\pm$0.33 & 87.20$\pm$0.26 & 87.77$\pm$0.83 & 83.40$\pm$1.20 & 87.08$\pm$0.65 & 74.13$\pm$3.90 \\
        ARFL         & 85.87$\pm$1.85 & 83.14$\pm$3.45 & 76.77$\pm$1.90 & 64.31$\pm$3.73 & 73.22$\pm$1.48 & 53.23$\pm$1.67 \\
        JointOpt     & 84.42$\pm$0.70 & 83.01$\pm$0.88 & 80.82$\pm$1.19 & 74.09$\pm$1.43 & 76.13$\pm$1.15 & 66.16$\pm$1.71 \\
        DivideMix    & 77.35$\pm$0.20 & 74.40$\pm$2.69 & 72.67$\pm$3.39 & 72.83$\pm$0.30 & 68.66$\pm$0.51 & 68.04$\pm$1.38 \\
        FedCorr      & 94.01$\pm$0.22 & 94.15$\pm$0.18 & 92.93$\pm$0.25 & 92.50$\pm$0.28 & 91.52$\pm$0.50 & 90.59$\pm$0.70 \\
        FedDiv (Ours)& \textbf{94.42$\pm$0.29} & \textbf{94.30$\pm$0.19} & \textbf{93.67$\pm$0.22} & \textbf{93.41$\pm$0.21} & \textbf{92.98$\pm$0.60} & \textbf{91.44$\pm$0.25} \\
        \bottomrule
    \end{tabular}%
    \caption{Best test accuracy (\%) of FedDiv and existing SOTA methods on CIFAR-10 with IID setting at diverse noise levels.}
    \label{Table:CIFAR10-IID}%
\end{table*}%

As shown in Figure~\ref{Figure:Framework}, the training procedure of the $t$-th communication round performs the following steps:

\hangindent 1.0em
\hangafter=0
    \noindent \textbf{Step\circled{1}}: 
    The server broadcasts the parameters of the global neural network model~$\theta^{(t)}$ and the federated noise filtering model~${\mathcal{W}^{(t)}}$ to every client~$k\in\mathcal{S}$, where $\mathcal{S}\subseteq\mathcal{M}$ is a subset of clients randomly selected with a fixed fraction $\omega={|\mathcal{S}|}/{K}$ in this round.

\hangindent 1.0em
\hangafter=0
    \noindent \textbf{Step\circled{2}}: 
        On every $k\in\mathcal{S}$, ${\mathcal{W}^{(t)}}$ is used to separate $\mathcal{D}_{k}$ into noisy and clean samples, and those noisy samples with high prediction confidence are assigned pseudo-labels predicted by $\theta^{(t)}$.
 
\hangindent 1.0em
\hangafter=0
    \noindent \textbf{Step\circled{3}}(Local model training): 
        Every client $k\in\mathcal{S}$ trains its local neural network using a subset of the clean and relabeled noisy samples selected using PCS to obtain its updated local parameters $\theta^{(t)}_{k}$. Here, we use $\hat{{\theta}}_k^{(t)}$ to denote the parameters of the local model being optimized during this training session.
 
\hangindent 1.0em
\hangafter=0
    \noindent \textbf{Step\circled{4}}(Local filter training): 
        Every client $k\in\mathcal{S}$ trains a local noise filter model with updated parameters $\mathcal{W}^{(t)}_{k}$ by fitting the per-sample loss function values of its private data $\mathcal{D}_{k}$. Such loss function values are evaluated using the logits of training samples in $\mathcal{D}_{k}$ predicted by $\theta^{(t)}_{k}$.

\hangindent 1.0em
\hangafter=0
    \noindent \textbf{Step\circled{5}}: 
        The server aggregates $\{{\theta}_{k}^{(t)}|k\in\mathcal{S}\}$ and then updates the global model as follows,
        \begin{equation}
            \label{Equation:FedAvg}
            \theta^{(t+1)} \leftarrow \sum_{k\in \mathcal{S}} \frac{n_k}{\sum_{k\in \mathcal{S}} n_k} {\theta}_{k}^{(t)}.
        \end{equation}

\hangindent 1.0em
\hangafter=0
    \noindent \textbf{Step\circled{6}}(Federated filter aggregation): 
        The server collects $\{{\mathcal{W}}_{k}^{(t)}|k\in\mathcal{S}\}$ to update existing server-cached local filter parameters $\{{\mathcal{W}}_{k}|k\in\mathcal{S}\}$, and then aggregates all local filters $\{{\mathcal{W}}_{k}|k\in\mathcal{M}\}$ to obtain an updated federated noise filtering model~${\mathcal{W}^{(t+1)}}$.

\noindent
This training procedure is repeated until the global model converges steadily or the pre-defined number of communication rounds~$\mathcal{T}$ is reached. Each communication round involves local training sessions (Step \circled{1}-\circled{4}) on several randomly selected clients and the model aggregation phase (Step \circled{5}-\circled{6}) on the server. The details of Step\circled{2}, Step\circled{4}, and Step\circled{6} are provided in ``Federated Noise Filter'', while the detailed description of Step\circled{3} is given in ``Predictive Consistency Based Sampler'' and ``Objectives for Local Model Training''.

\subsection{Federated Noise Filter}
 
Aiming at identifying label noise for the F-LNL task, we propose a Federated Noise Filter (FNF), which models the global distribution of clean and noisy samples across all clients. Motivated by~\cite{fedgmm2022}, this FNF model can be constructed via the federated EM algorithm. Specifically, we first conduct local filter training to obtain locally estimated GMM parameters. This goal is reached by iteratively executing the standard EM algorithm~\cite{GMM_EM} per client to fit its local noise distribution. Then, we perform federated filter aggregation to aggregate local GMM parameters received from all clients to construct a federated noise filter. 

\noindent
\textbf{Local filter training.} 
In general, samples with label noise tend to possess higher loss function values during model training, making it feasible to use mixture models to separate noisy samples from clean ones using per-sample loss values of the training samples~\cite{BMM, li2019dividemix}. 
Therefore, for the $k$-th client in the $t$-th communication round, 
a local GMM can be built to model the local distribution of clean and noisy samples by fitting the per-sample loss distribution,
\begin{equation}
    \{\ell(x,y;{\theta}_{k}^{(t)})|(x,y)\in\mathcal{D}_{k}\},
\end{equation}
where $\ell(x,y;{\theta}_{k}^{(t)})$ is the cross-entropy loss of a sample-label pair $(x, y)\in\mathcal{D}_{k}$ when the local model ${\theta}_{k}^{(t)}$ is used for prediction. Then, we denote the two-component GMM model by $\mathcal{W}_{k}^{(t)}=(\bm{{\mu}}_{k}^{(t)},\bm{{\sigma}}_{k}^{(t)},{\bm{\pi}}_{k}^{(t)})$, 
where $\bm{{\mu}}_{k}^{(t)}$ and $\bm{{\sigma}}_{k}^{(t)}$ are vectors with entries ${{\mu}}_{kg}^{(t)}$ and ${\sigma}_{kg}^{(t)}$ denoting the mean and variance of the $g$-th Gaussian component, respectively. Here, we set $g=1$ to represent the ``clean'' Gaussian component, i.e., the Gaussian component with a smaller mean (smaller loss), while $g=2$ denotes the ``noisy'' one.

We further define a discrete variable $z$ to represent whether a sample is clean or noisy. ${\bm{\pi}}_{k}^{(t)}$ denotes the prior distribution of $z$, i.e., ${{\pi}}_{kg}^{(t)}=P(\!z\!=\!g\!)$, where ${{\pi}}_{kg}^{(t)}$ should satisfy  $\sum_{g=1}^{2}{{\pi}}_{kg}^{(t)}=1$ and $0 \le{{\pi}}_{kg}^{(t)}\le 1$ for $g=1, 2$. 
Thus, $P(\ell(x,y;{\theta}_{k}^{(t)})|z=g)$ is modeled as a Gaussian distribution $\mathcal{N}(\ell(x,y;{\theta}_{k}^{(t)}); \mu_{kg}^{(t)}, \sigma_{kg}^{(t)})$.
Then, the posterior ${\gamma}_{kg}(x,y;{\theta}_{k}^{(t)})$, which represents the probability of a sample $x$ being clean ($g=1$) or noisy ($g=2$) given its loss value, can be computed as
 \begin{equation}
    \label{Equation:E-step}
    \begin{split}
        {\gamma}_{kg}&(x,y;{\theta}_{k}^{(t)}) = P(z=g|x,y;{\theta}_{k}^{(t)})\\ &= \frac{P(\ell(x,y;{\theta}_{k}^{(t)})|z=g) P(z=g)}{\sum_{g^{\prime}=1}^{2} P(\ell(x,y;{\theta}_{k}^{(t)})|z=g^{\prime}) P(z=g^{\prime})}.
    \end{split}
\end{equation}

Afterwards, for each client ${k}$, leveraging its private data $\mathcal{D}_{k}$, updated local parameters $\theta^{(t)}_{k}$, and the federated filter parameters $\mathcal{W}^{(t)}$ received from the server, its optimal local filter parameters $\mathcal{W}_{k}^{(t)}=(\boldsymbol{\mu}_{k}^{(t)}, \boldsymbol{\sigma}_{k}^{(t)}, \boldsymbol{\pi}_{k}^{(t)})$ at current round $t$ are derived through the training of the local GMM models utilizing a standard EM algorithm~\cite{GMM_EM}. Noted that, $\mathcal{W}^{(t)}$ originated from the server here is used to initialize $\mathcal{W}_{k}^{(t)}$ for expediting convergence.

In the $t$-th communication round, once we have performed local filter training on the clients in $\mathcal{S}$, we upload the local filter parameters $\{\mathcal{W}_{k}^{(t)}|k\in\mathcal{S}\}$ to the server. Then, the server updates its cached local filter parameters corresponding to each client in $\mathcal{S}$ as follows,
\begin{equation}
    \begin{split}
        \{{\mathcal{W}}_{k} \leftarrow {\mathcal{W}}_{k}^{(t)} | k\in\mathcal{S}\},
    \end{split}
\end{equation}
where ${\mathcal{W}}_{k}$ is the server-cached version of the local noise filter on the $k$-th client. 
Note that \texttt{FedDiv} only sends three numerical matrices (i.e., $\boldsymbol{\mu}_{k}^{(t)}$, $\boldsymbol{\sigma}_{k}^{(t)}$, and $\boldsymbol{\pi}_{k}^{(t)}$) from each client to the server, and they merely reflect each client's local noise distributions instead of the raw input data, avoiding any risk of data privacy leakage.

\begin{table}[t]
    \centering
    \small
    \begin{tabular}{l|ccc}
        \toprule
        \midrule
        \multicolumn{1}{c|}{\textbf{Method}} & \multicolumn{3}{c}{\textbf{Best Test Accuracy $\pm$ Standard Deviation}} \\
        & $p=0.7$ & $p=0.7$ & $p=0.3$ \\
        & $\alpha_{Dir}=10$ & $\alpha_{Dir}=1$ & $\alpha_{Dir}=10$ \\
        \midrule
        FedAvg       & 78.88$\pm$2.34 & 75.98$\pm$2.92 & 67.75$\pm$4.38 \\
        RoFL         & 79.56$\pm$1.39 & 72.75$\pm$2.21 & 60.72$\pm$3.23 \\
        ARFL         & 60.19$\pm$3.33 & 55.86$\pm$3.30 & 45.78$\pm$2.84 \\
        JointOpt     & 72.19$\pm$1.59 & 66.92$\pm$1.89 & 58.08$\pm$2.18 \\
        DivideMix    & 65.70$\pm$0.35 & 61.68$\pm$0.56 & 56.67$\pm$1.73 \\
        FedCorr      & 90.52$\pm$0.89 & 88.03$\pm$1.08 & 81.57$\pm$3.68 \\
        FedDiv (Ours)& \textbf{93.18$\pm$0.42} & \textbf{91.95$\pm$0.26} & \textbf{85.31$\pm$2.28} \\
        \bottomrule
    \end{tabular}
    \caption{Best test accuracy (\%) of FedDiv and existing SOTA methods on CIFAR-10 with non-IID setting at the noise level $(\rho, \tau)=(6.0, 0.5)$.}
    \label{Table:CIFAR10-NonIID}%
\end{table}

\noindent
\textbf{Federated filter aggregation.} 
After parameter uploading, the federated filter model can be constructed by aggregating the local filter parameters corresponding to all the clients $\{\mathcal{W}_{k}=(\bm{\mu}_{k}, \bm{\sigma}_{k}, \bm{{\pi}}_{k})|k\in\mathcal{M}\}$  as follows, 
\begin{equation}
    \label{Equation:M-Step-at-Server}
    \begin{split}
        {{{\mu}}^{(t+1)}_{g}} &= \sum_{k\in \mathcal{M}} \frac{n_k}{\sum_{k\in \mathcal{M}} n_k}  {{\mu}}_{kg}, \\
        {{\sigma}^{(t+1)}_{g}} &= \sum_{k\in \mathcal{M}} \frac{n_k}{\sum_{k\in \mathcal{M}} n_k} {{\sigma}}_{kg}, \\
        {{{\pi}}^{(t+1)}_{g}} &= \sum_{k\in \mathcal{M}} \frac{n_k}{\sum_{k\in \mathcal{M}} n_k} {{\pi}}_{kg},
    \end{split}
\end{equation}
where $\mathcal{W}^{(t+1)}=(\bm{\mu}^{(t+1)}, \bm{\sigma}^{(t+1)}, \bm{{\pi}}^{(t+1)})$ will be used to perform label noise filtering on the selected clients at the beginning of the $(t+1)$-th communication round.

\noindent
\textbf{Label noise filtering.}
In the $t$-th communication round, once the $k$-th client has received the parameters of the global model~$\theta^{(t)}$ and the federated filter model~$\mathcal{W}^{(t)}$ from the server, the probability of a sample $x$ from $\mathcal{D}_k$ being clean can be estimated through its posterior probability for the ``clean'' component as follows,
\begin{equation}
    \begin{split}
        \textbf{\text{p}}(&\text{``clean''}|x,y;{\theta}^{(t)})=P(z=1|x,y;{\theta}^{(t)}).
    \end{split}
\end{equation}
Afterwards, we can divide the samples of $\mathcal{D}_{k}$ into a clean subset $\mathcal{D}_{k}^{\text{clean}}$ and a noisy subset $\mathcal{D}_{k}^{\text{noisy}}$ by thresholding their probabilities of being clean with the threshold $0.5$ as follows,
\noindent
\begin{equation}
    \label{Equation:Label-Splitting}
    \begin{split}
        \mathcal{D}_{k}^{\text{clean}} \leftarrow \{\left(x, y\right) | \textbf{\text{p}}(&\text{``clean''}|x,y;{\theta}_{k}^{(t)})\ge0.5, \forall \left(x, y\right)\in\mathcal{D}_{k}\}, \\
        \mathcal{D}_{k}^{\text{noisy}} \leftarrow \{\left(x, \right) | \textbf{\text{p}}(&\text{``clean''}|x,y;{\theta}_{k}^{(t)})<0.5, \forall \left(x, y\right)\in\mathcal{D}_{k}\}.
    \end{split}
\end{equation}

\noindent
\textbf{Noisy sample relabeling.}
We compute the noise level of the $k$-th client as $\hat{\delta}_{k} = |\mathcal{D}_{k}^{\text{noisy}}|/|\mathcal{D}_{k}|$, while a client is considered a noisy one if $\hat{\delta}_{k} > 0.1$; and otherwise, a clean one.
For an identified noisy client, we simply discard the given labels of noisy samples from $\mathcal{D}_{k}^{\text{noisy}}$ to prevent the model from memorizing label noise in further local training. In an effort to leverage these unlabeled (noisy) samples, we relabel those samples with high prediction confidence (by setting a confidence threshold $\zeta$) by assigning predicted labels from the global model as follows,
\begin{equation}
    \label{Equation:Relabel}
    \mathcal{D}_{k}^{\text{relab}} \leftarrow \{\left(x, \hat{y}\right) | \max(\textbf{\text{p}}(x;{\theta}^{(t)})) \ge \zeta, \forall x\in \mathcal{D}_{k}^{\text{noisy}}\},
\end{equation}
where $\hat{y}= \hat{y}(x)= \argmax \textbf{\text{p}}(x;{\theta}^{(t)})$ is the pseudo-label for the sample $x$ predicted by the global model ${\theta}^{(t)}$.

\begin{table}[t!]
    \centering
    \small
    \begin{tabular}{l|ccc}
        \toprule
        \midrule
        \multicolumn{1}{c|}{\multirow{3}[2]{*}{\textbf{Method}}} & \multicolumn{3}{c}{\textbf{Best Test Accuracy $\pm$ Standard Deviation}} \\
        & $\rho$=0.4 & $\rho$=0.6 & $\rho$=0.8 \\
        & $\tau$=0.5 & $\tau$=0.5 & $\tau$=0.5 \\
        \midrule
        FedAvg       & 64.41$\pm$1.79 & 53.51$\pm$2.85 & 44.45$\pm$2.86\\ 
        RoFL         & 59.42$\pm$2.69 & 46.24$\pm$3.59 & 36.65$\pm$3.36\\
        ARFL         & 51.53$\pm$4.38 & 33.03$\pm$1.81 & 27.47$\pm$1.08\\
        JointOpt     & 58.43$\pm$1.88 & 44.54$\pm$2.87 & 35.25$\pm$3.02\\
        DivideMix    & 43.25$\pm$1.01 & 40.72$\pm$1.41 & 38.91$\pm$1.25\\
        FedCorr      & 74.43$\pm$0.72 & 66.78$\pm$4.65 & 59.10$\pm$5.12\\
        FedDiv (Ours)& \textbf{74.86$\pm$0.91} & \textbf{72.37$\pm$1.12} & \textbf{65.49$\pm$2.20}\\
        \bottomrule
    \end{tabular}
    \caption{Best test accuracy (\%) of FedDiv and existing SOTA methods on CIFAR-100 with IID setting at diverse noise levels.}
    \label{Table:CIFAR100-IID}
\end{table}%

\subsection{Predictive Consistency Based Sampler}

During the $t$-th round on client $k$, upon obtaining the clean subset $\mathcal{D}{k}^{\text{clean}}$ and the relabeled subset $\mathcal{D}{k}^{\text{relab}}$, we integrate them into supervised local model training across $T$ local epochs. However, the complete elimination of label noise among clients during noise filtering and relabeling is unattainable. On the other hand, relabeling inevitably introduces new label noise, causing instability in local model training, which further negatively affects the performance of the global model during aggregation. To tackle this, we propose a Predictive Consistency based Sampler (PCS) to reselect labeled samples for local training. Specifically, we observe enforcing the consistency of class labels respectively predicted by global and local models is a good strategy to achieve this goal. As training proceeds, the robustness of the model against label noise would be significantly increased (See below.), thus easily improving predictions' reliability of the samples having new label noise. 

In addition, due to data heterogeneity in federated settings, especially for non-IID data partitions, local training samples owned by individual clients often belong to a smaller set of dominant classes. Thus, samples of dominant classes with newly introduced label noise would be better self-corrected during local model training, gradually leading to inconsistent predictions w.r.t those of the global model.
However, such class-unbalanced local data would also contribute to the cause of the local model bias towards the dominant classes~\cite{imbalance}, which makes it more difficult for the local model to produce correct pseudo-labels for the samples that belong to minority classes. The proposed PCS strategy mitigates model bias to improve the reliability of class labels produced by local models. Here, we can de-bias the model predictions by improving causality via counterfactual reasoning~\cite{holland1986statistics, pearl2009causal, debiasedlearning}, and therefore, the de-biased logit of a sample $x$ from $\mathcal{D}_{k}^{\text{clean}}$ or $\mathcal{D}_{k}^{\text{relab}}$ is induced as follows,
\begin{equation}
    \label{Equation:Logit-Debias}
     F(x) \leftarrow f(x;\hat{{\theta}}_k^{(t)}) -\xi  \text{log} (\hat{p}_{k}),
\end{equation}
where $F(x)$ is the de-biased logit later used for generating the de-biased pseudo-label, i.e., $\tilde{y}(x) = \argmax F(x)$, and $\xi =0.5$ is a de-bias factor. $f(x;\hat{{\theta}}_k^{(t)})$ is the original logit for the sample $x$ produced by the local model $\hat{{\theta}}_k^{(t)}$, currently being optimized. $\hat{p}_{k}$ represents the overall bias of the local model w.r.t all classes, which was previously updated according to Eq.~(\ref{Equation:Momentum-p}) and cached on the $k$-th client during the last local training session.
 
\begin{table}[t]
    \centering
    \small
    \begin{tabular}{l|cccc}
        \toprule
        \midrule
        \textbf{Dataset} & \multicolumn{1}{c}{\textbf{CIFAR-100}} & \multicolumn{1}{c}{\textbf{Clothing1M}} \\
        \textbf{Noise level ($\rho$, $\tau$)} & (0.4, 0.0) & - \\
        \textbf{Method\textbackslash ($p$, $\alpha_{Dir}$) }& (0.7, 10) & - \\
        \midrule
        FedAvg                   & 64.75$\pm$1.75 &  70.49 \\
        RoFL                     & 59.31$\pm$4.14 &  70.39 \\
        ARFL                     & 48.03$\pm$4.39 &  70.91 \\
        JointOpt                 & 59.84$\pm$1.99 &  71.78 \\
        DivideMix                & 39.76$\pm$1.18 &  68.83 \\
        FedCorr                  & 72.73$\pm$1.02 &  72.55 \\
        FedDiv (Ours)            & \textbf{74.47$\pm$0.34} &  \textbf{72.96$\pm$0.43} \\
        \bottomrule
    \end{tabular}%
    \caption{Best test accuracy (\%) of FedDiv and existing SOTA methods on CIFAR-100 and Clothing1M under the non-IID data partitions.}
    \label{Table:Clothing-1M}%
\end{table}%

Afterwards, PCS is used to re-select higher-quality and more reliable labeled training samples to perform local training as follows, 
\begin{equation}
    \label{Equation:Reselect}
    \begin{split}
        ~\mathcal{D}_{k}^{\text{resel}} \leftarrow \{\left(x, y\right) | \hat{y}(x)=\tilde{y}(x), \forall (x, y)\in \mathcal{D}_{k}^{\text{clean}} \cup \mathcal{D}_{k}^{\text{relab}}\}.
    \end{split}
\end{equation}

Similar to~\cite{xu2022fedcorr}, we update the training dataset for further optimizing the local model as follows,
\begin{equation}
    \label{Equation:Data-Split}
    \hat{\mathcal{D}}_{k} =
        \begin{cases} 
            \mathcal{D}_{k}^{\text{resel}}, &  \text{if}~\hat{\delta}_{k} \ge 0.1, \\
            \mathcal{D}_{k},  & \text{if}~\hat{\delta}_{k} < 0.1.
        \end{cases}
\end{equation}

Once having the optimized $\theta^{(t)}_k$ from a local training session, we use it to update $\hat{p}_{k}$ with momentum as follows,
\begin{equation}
    \label{Equation:Momentum-p}
    \hat{p}^{(t)}_{k} \leftarrow m\hat{p}_{k} +  (1-m)\frac{1}{n_{k}}\sum_{x\in \mathcal{D}_{k}} \textbf{\text{p}}(x;{\theta}^{(t)}_{k}),
\end{equation}
where $m=0.2$ is a momentum coefficient. We save $\hat{p}^{(t)}_{k}$ on the $k$-th client to update existing client-cached $\hat{p}_{k}$.

\subsection{Objectives for Local Model Training}

To enhance the model's robustness against label noise, we here use MixUp regularization~\cite{zhang2018mixup} for local model training, further undermining the instability of training. Specifically, two sample-label pairs $(x_i, y_i)$ and $(x_j, y_j)$ from $\hat{\mathcal{D}}_{k}$ are augmented using linear interpolation, ${\ddot{x}}={{\lambda}}{x_i}+({1-{\lambda}}){x_j}$ and ${\ddot{y}}={{\lambda}}{p_y(y_i)}+({1-{\lambda}}){p_y(y_j)}$, where ${\lambda}\sim Beta(\alpha)$ is a mixup ratio, $\alpha=1$ is a scalar to control its distribution, and $p_y(\cdot)$ is a function to generate a one-hot vector for a given label. Hence, on the $k$-th client in the $t$-th communication round, the local model $\hat{{\theta}}^{(t)}_k$ is trained with the cross-entropy loss applied to $B$ augmented samples in one mini-batch as follows,
\begin{equation}
    {\mathcal{L}}_{mix}=-\sum_{b=1}^{B}\ddot{y}_b\log{\textbf{\text{p}}(\ddot{x}_b;\hat{{\theta}}^{(t)}_k)}.
\end{equation}
 
With the high heterogeneity of the non-IID data partitions, there could be only a limited number of categories on each client, and extensive experiments have shown that such a data partition forces a local model to predict the same class label to minimize the training loss. As in \cite{tanaka2018joint, arazo2019unsupervised}, regularizing the average prediction of a local model over every mini-batch using a uniform prior distribution is a viable solution to overcome the above problem, i.e.,
\begin{equation}
    \mathcal{L}_{reg} = \sum_{c=1}^{C} \hat{\textbf{\text{q}}}^{c} \log \left(\frac{\hat{\textbf{\text{q}}}^{c}}{\textbf{\text{q}}^{c}}\right), \text{ where } \textbf{\text{q}}=\frac{1}{B} \sum_{b=1}^{B} \textbf{\text{p}}(\ddot{x}_b;\hat{{\theta}}^{(t)}_k),
\end{equation}
where $\hat{\textbf{\text{q}}}^{c}={1}/{C}$ denotes the prior probability of a class $c$. $\textbf{\text{q}}^c$ is the $c$-th element of the vector $\textbf{\text{q}}$, which refers to the predicted probability of the class $c$ averaged over $B$ augmented training samples in a mini-batch.

Finally, on the $k$-th client in the $t$-th communication round, the overall loss function for optimizing the local model is defined as follows,
\begin{equation}
    \label{Equation:Final-Loss}
    \mathcal{L}_{final} = \mathcal{L}_{mix} + \eta \mathcal{L}_{reg},
\end{equation}
where $\eta$ is a weighting factor balancing $\mathcal{L}_{mix}$ and $\mathcal{L}_{reg}$. In the experiments, we set $\eta=1$ when the data partition is non-IID; and otherwise, $\eta=0$.

\section{Experiments}

\subsection{Experimental Setups}

To be fair, we here adopt the consistent experimental setups with \texttt{FedCorr}~\cite{xu2022fedcorr} to assess the efficacy of our proposed approach \texttt{FedDiv}. Further details, e.g., data partitions and additional implementations and analysis, are provided in the supplementary document (abbr. \textbf{Supp}\footnote{https://github.com/lijichang/FLNL-FedDiv/blob/main/supp.pdf}).

\noindent
\textbf{Datasets and data partitions.} 
We validate \texttt{FedDiv}'s superiority on three classic benchmark datasets, including two synthetic datasets namely CIFAR-10~\cite{cifar} and CIFAR-100~\cite{cifar}, and one real-world noisy dataset, i.e., Clothing1M~\cite{clothing1m}.
Like~\cite{xu2022fedcorr}, we take both IID and non-IID data partitions into account on CIFAR-10 and CIFAR-100, but only consider non-IID data partitions on Clothing1M. Under the IID data partitions, each client is randomly assigned the same number of samples with respect to each class. For non-IID data partitions, it is constructed using a Dirichlet distribution~\cite{lin2020ensemble} with two pre-defined parameters, namely the fixed probability $p$ and the concentration parameter $\alpha_{Dir}$.

\noindent
\textbf{Label noise settings.}
Similar to~\cite{xu2022fedcorr}, the noise level for the $k$-th client can be defined as follows:
\begin{equation}
    \label{Equation:Noise}
    {\delta}_{k} =
    \begin{cases} 
        u \sim U(\tau, 1),  & \text{probability of}~\rho, \\
        0, &  \text{probability of}~1-\rho.
    \end{cases}
\end{equation}
Here, $\rho$ signifies the probability of a client being noisy. For a noisy client with ${\delta}_{k}\ne0$, the noise level is initially sampled at random from the uniform distribution $u \sim U(\tau, 1)$, with $\tau$ being its lower bound. Subsequently, ${\delta}_{k} \cdot 100\%$ of local examples are randomly selected as noisy samples, with their ground-truth labels replaced by all possible class labels.

\begin{figure}[t]
    \centering
    \includegraphics[width=0.45\textwidth]{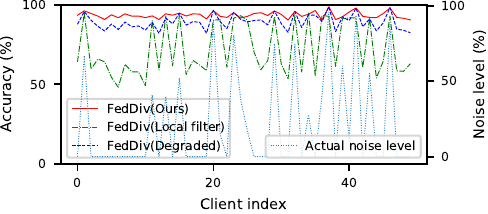}
    \caption{The accuracy of noisy label identification across different clients, with the experiment conducted on CIFAR-100 with the non-IID data partition under the noise setting $(\rho, \tau)=(0.4, 0.0)$. (Best viewed zoomed in.)} 
    \label{Figure:Filtering-Accuracy-per-Method}
\end{figure}

\noindent
{\bf Baselines.}
We compare \texttt{FedDiv} with existing state-of-the-art (SOTA) F-LNL methods, including \texttt{FedAvg}~\cite{FedAvg}, \texttt{RoFL}~\cite{RoFL}, \texttt{ARFL}~\cite{RoFL}, \texttt{JointOpt}~\cite{JointOpt}, \texttt{DivideMix}~\cite{li2019dividemix}, and \texttt{FedCorr}~\cite{xu2022fedcorr}. Their experimental results reported in this paper are borrowed from~\cite{xu2022fedcorr}. 
 
\noindent
\textbf{Implementations.}
We set $\omega$ and $\mathcal{T}$ to 950, 900 and 100, and 0.1, 0.1 and 0.02 on CIFAR-10, CIFAR-100 and Clothing1M, respectively, while we also set the confidence threshold $\zeta=0.75$ for relabeling on all datasets. Note that, to enable faster convergence, we warm up local neural network models of each client for $\mathcal{T}_{wu}$ iterations (not training rounds; see~\cite{xu2022fedcorr}) using MixUp regularization~\cite{zhang2018mixup}. Additionally, to be fair, most of our implementation details involving both local training and model aggregation are consistent with \texttt{FedCorr}~\cite{xu2022fedcorr} for each dataset under all federated settings and all label noise settings.

\noindent
\textbf{Model variants.}
We build the variants of~\texttt{FedDiv} to evaluate the effect of the proposed noise filter as follows.

\begin{itemize}
    \item \texttt{FedDiv(Degraded)}: Following~\cite{fedgmm2022}, we here degrade the proposed federated noise filter by constructing the global noise filter using only the local filter parameters received in the current round instead of those from all clients.
    \item \texttt{FedDiv(Local filter)}: A local noise filter is trained for each client using its own private data to identify noisy labels within individual clients.
\end{itemize}

\subsection{Comparisons with State-of-the-Arts}

Tables~\ref{Table:CIFAR10-IID}-\ref{Table:Clothing-1M} summarize classification performance of \texttt{FedDiv} against state-of-the-art (SOTA) F-LNL methods on CIFAR-10, CIFAR-100, and Clothing1M across various noise settings for both IID and non-IID data partitions. Comparison results, based on mean accuracy and standard deviation over five trials, demonstrate \texttt{FedDiv}'s significant superiority over existing F-LNL algorithms, especially in challenging cases. For instance, in IID data partitions, Table~\ref{Table:CIFAR100-IID} illustrates \texttt{FedDiv} outperforming \texttt{FedCorr} on CIFAR-100 by 6.39\% in the toughest noise setting with $(\rho, \tau)=(0.8, 0.5)$. Similarly, for non-IID partitions in Table~\ref{Table:CIFAR10-NonIID}, \texttt{FedDiv} consistently surpasses \texttt{FedCorr} by 3.74\% in the most challenging setting $(p, \alpha_{Dir})=(0.3, 10)$ on CIFAR-10. Additionally, in Table~\ref{Table:Clothing-1M}, \texttt{FedDiv} exhibits a 0.41\% improvement over \texttt{FedCorr} on Clothing1M, indicating its efficacy in real-world label noise distributions.

\begin{table}[t]
    \centering
    \small
    \begin{tabular}{l|cccc}
        \toprule
        \midrule
        \textbf{Dataset} & \multicolumn{1}{c}{\textbf{CIFAR-10}} & \multicolumn{1}{c}{\textbf{CIFAR-100}} \\
        \textbf{Noise level ($\rho$, $\tau$)} & (0.6, 0.5) & (0.4,0.0) \\
        \textbf{Method\textbackslash ($p$, $\alpha_{Dir}$) }& (0.3, 10) & (0.7, 10) \\
        \midrule
        FedDiv (Ours)                  & 85.31$\pm$2.28 &  74.47$\pm$0.34 \\
        \midrule
        FedDiv (Degraded)              & 83.22$\pm$2.61 &  73.06$\pm$0.93 \\
        FedDiv (Local filter)          & 81.34$\pm$3.65 &  71.37$\pm$0.76 \\
        \midrule
        FedDiv w/o Relab. \& w/o PCS   & 82.17$\pm$3.06 &  73.09$\pm$1.89 \\
        FedDiv w/o PCS                 & 82.83$\pm$2.59 &  73.66$\pm$0.96 \\
        FedDiv w/o $\mathcal{L}_{reg}$ & 83.60$\pm$3.65 &  72.43$\pm$1.29 \\
        \bottomrule
    \end{tabular}%
    \caption{Ablation study results for CIFAR-10 and CIFAR-100, with varying noise levels under non-IID data partitions.}
    \label{Table:Ablation}%
\end{table}%

\subsection{Ablation Analysis}

To underscore the efficacy of \texttt{FedDiv}, we perform an ablation study to demonstrate the effect of each component.

\noindent
\textbf{Evaluation of federated noise filtering.}
To affirm the superiority of the proposed scheme for label noise filtering, we first compare \texttt{FedDiv} with our model variants \texttt{FedDiv(Local filter)} and \texttt{FedDiv(Degraded)}. As per Figure~{\ref{Figure:Filtering-Accuracy-per-Method}} and Table~{\ref{Table:Ablation}}, the proposed noise filter exhibits a superior capacity of identifying label noise on both clean and noisy clients, leading to considerably improved classification performance in comparison to its two variants.
 
\noindent
\textbf{Evaluation of relabeling and re-selection.} 
To assess the efficacy of the proposed strategies for noisy sample relabeling and labeled sample re-selection, we systematically remove their respective components from the \texttt{FedDiv} framework.  The results depicted in Table~{\ref{Table:Ablation}} demonstrate a substantial decrease in accuracy across various noise settings for both types of data partitions. This indicates the importance of each individual component. 

\section{Conclusions}

In this paper, we have presented \texttt{FedDiv} to handle the task of Federated Learning with Noisy Labels (F-LNL). It can effectively respond to the challenges in F-LNL tasks involving both data heterogeneity and noise heterogeneity while taking privacy concerns into account. On the basis of an FL framework, we first propose Federated Noise Filtering to separate clean samples from noisy ones on each client, thereby diminishing the instability during training. Then we perform relabeling to assign pseudo-labels to noisy samples with high predicted confidence. In addition, we introduce a Predictive Consistency based Sampler to identify credible local data for local model training, thus avoiding label noise memorization and further improving training stability. Experiments as well as comprehensive ablation analysis have revealed \texttt{FedDiv}'s superiority in handling F-LNL tasks.

\section{Acknowledgements}

This work was supported in part by the National Natural Science Foundation of China (NO.~62322608), in part by the Shenzhen Science and Technology Program (NO.~JCYJ20220530141211024), in part by the Open Project Program of the Key Laboratory of Artificial Intelligence for Perception and Understanding, Liaoning Province (AIPU, No.~20230003), in part by Hong Kong Research Grants Council under Collaborative Research Fund (Project No.~HKU~C7004-22G).

\bibliography{aaai24}

\clearpage

\label{supp}

\begin{strip}
\centering
\textbf{\Large Supplementary Material (abbr. Supp)}
\end{strip}

In this supplementary material, we detail additional experimental setups encompassing data partitions, implementation specifics, and the baseline F-LNL methods. Furthermore, we provide further analysis of our proposed method, \texttt{FedDiv}. For a comprehensive understanding, we summarize the key notations of F-LNL and \texttt{FedDiv} in Table~\textcolor{red}{\ref{Table:Notations}}. Additionally, detailed training procedures of \texttt{FedDiv} and the specifics of local filter training are outlined in Algorithm~\textcolor{red}{\ref{PseudoCode:FedDiv}} and Algorithm~\textcolor{red}{\ref{PseudoCode:Local-Model-Training}}, respectively. To further enhance clarity, we present hyper-parameter summaries for each dataset in Table~\textcolor{red}{\ref{Table:Hyper-Parameter}}. Consistency in hyper-parameter settings is maintained across different label noise settings for both IID and non-IID data partitions on each dataset. It's worth noting that all experiments are conducted on the widely-used PyTorch platform\footnote{https://pytorch.org/} and executed on an NVIDIA GeForce GTX 2080Ti GPU with 12GB memory.

\begin{table}[t]
    \centering
    \begin{tabular}{lp{0.65\linewidth}}
        \toprule
        \midrule
        \textbf{Symbol} & \textbf{Definition} \\
        \midrule
        $t$                      & Current communication round \\
        $\mathcal{M}$            & Collection of all clients \\
        $\mathcal{S}$            & Collection of clients randomly selected at current round \\
        $k$                      & Current index for the client selected from $\mathcal{M}$ or $\mathcal{S}$ \\
        $\mathcal{D}_{k}$        & Given local samples for client $k$ \\
        $\mathcal{D}_{k}^{\text{clean}}$ & Clean samples separated from $\mathcal{D}_{k}$ \\
        $\mathcal{D}_{k}^{\text{noisy}}$ & Noisy samples separated from $\mathcal{D}_{k}$ \\
        $\mathcal{D}_{k}^{\text{relab}}$ & Relabeled samples produced from $\mathcal{D}_{k}^{\text{noisy}}$ \\
        $\mathcal{D}_{k}^{\text{resel}}$  & More reliable labeled samples re-selected by PCS \\
        $\hat{\mathcal{D}}_{k}$  & Training data for further optimization of the local model \\
        $\delta_{k}$             & Actual noise level for client $k$ \\
        $\hat{\delta}_{k}$       & Estimated noise level for client $k$ \\
        $\theta^{(t)}$           & Global network model at round $t$\\
        $\hat{\theta}_{k}^{(t)}$ & Local network model being optimized during local training for client $k$ at round $t$ \\
        $\theta_{k}^{(t)}$       & Local network model for client $k$ at round $t$\\
        $\mathcal{W}^{(t)}$      & Global filter parameters for client $k$ at round $t$ \\
        $\mathcal{W}_{k}$        & Server-cached local filter parameters for client $k$\\
        $\mathcal{W}_{k}^{(t)}$  & Local filter parameters for client $k$ obtained at round $t$\\
        $\hat{p}_{k}$            & Overall bias of the local model for client $k$ w.r.t all classes \\
        \bottomrule
    \end{tabular}%
    \caption{Key notations of F-LNL and FedDiv.}
    \label{Table:Notations}%
\end{table}%

\begin{table}[t]
    \centering
    \resizebox{\linewidth}{!}{
        \begin{tabular}{lccc}
            \toprule
            \midrule
            \textbf{Hyper-parameter} & \textbf{CIFAR-10} & \textbf{CIFAR-100} & \textbf{Clothing1M} \\
            \midrule
            \# of clients ($K$)      & 100       & 50        & 500                   \\
            \# of classes ($C$)      & 10        & 100       & 14                    \\
            \# of samples            & 50,000    & 50,000    & 1,000,000             \\
            Architecture             & ResNet-18 & ResNet-34 & Pre-trained ResNet-50 \\
            Mini-batch size          & 10        & 10        & 16                    \\
            Learning rate            & 0.03      & 0.01      & 0.001                 \\
            $\mathcal{T}_{wu}$       & 5         & 10        & 2                     \\
            $\mathcal{T}_{ft}$       & 500       & 450       & 50                    \\
            $\mathcal{T}_{ut}$       & 450       & 450       & 50                    \\
            $\mathcal{T}$            & 950       & 900       & 100                   \\
            $T$                      & 5         & 5         & 5                     \\
            $\omega$                 & 0.1       & 0.1       & 0.02                  \\
            \bottomrule
        \end{tabular}%
    }

    \caption{Hyper-parameters on different datasets.}
    \label{Table:Hyper-Parameter}%
\end{table}%

\begin{algorithm}[t]
    \small

    \textbf{Input}: {$\mathcal{M}$; $\{\mathcal{D}_{k}|k\in\mathcal{M}\}$; $\mathcal{T}$; $T$;  $\omega$; Initialized $\theta^{(0)}$; Initialized $\mathcal{W}^{(0)}$; Initialized  $\{\mathcal{W}_{k}|k\in\mathcal{M}\}$; Initialized $\{\hat{p}_{k}|k\in\mathcal{M}\}$}\\
    \textbf{Output}: {Global network model $\theta^{(\mathcal{T})}$}
      
    \begin{algorithmic}[1]
        \FOR{$k \in \mathcal{M}$}{
            \STATE $[k] \leftarrow (\mathcal{D}_{k}, \hat{p}_{k})$ \hfill{\textcolor{blue}{$\triangleright$  Packaging}}
        }\ENDFOR

        \item[] \textcolor{blue}{// Model warming-up step}
        
        \FOR{$t = 1$ to $\mathcal{T}_{wu}$}{
        
            \STATE Warm up $\theta^{(t)}$ for each client $k\in\mathcal{M}$
            
        }\ENDFOR
        
        \item[] {\textcolor{blue}{// Model training step}}
        
        \FOR{$t$ = $1$ to $\mathcal{T}$}{
        
        \STATE $\mathcal{S} \leftarrow$ Randomly select $\omega\times100\%$ clients from $\mathcal{M}$ 
        
        \FOR{$k \in \mathcal{S}$}{
        
            \STATE $(\mathcal{D}_{k}, \hat{p}_{k})  \leftarrow [k]$ \hfill{\textcolor{blue}{$\triangleright$ Unpackaging}}
            
            \STATE Obtain $\mathcal{D}_{k}^{\text{clean}}$ and $\mathcal{D}_{k}^{\text{noisy}}$ via the federated noise filter model ${\mathcal{W}}^{(t)}$ using Eq.~(\textcolor{red}{\ref{Equation:Label-Splitting}})\;
            
            \STATE Estimate $\hat{\delta}_{k} = |\mathcal{D}_{k}^{\text{noisy}}|/|\mathcal{D}_{k}|$\;
            
            \STATE Obtain $\mathcal{D}_{k}^{\text{relab}}$ using Eq.~(\textcolor{red}{\ref{Equation:Relabel}})\;
            
            \STATE $\hat{\theta}_{k}^{(t)} \leftarrow \theta^{(t)}$\;
            
            \FOR{$t^{\prime} = 1$  to $T$}{
            
                \STATE Obtain $\mathcal{D}_{k}^{\text{opt}}$ using Eqs.~(\textcolor{red}{\ref{Equation:Logit-Debias}}), (\textcolor{red}{\ref{Equation:Reselect}}), (\textcolor{red}{\ref{Equation:Data-Split}})\;
                
                \STATE Optimize $\hat{\theta}_{k}^{(t)}$ using  Eq.~(\textcolor{red}{\ref{Equation:Final-Loss}})\;
                
            }\ENDFOR
            
            \STATE $\theta_{k}^{(t)} \leftarrow \hat{\theta}_{k}^{(t)}$\;
            
            \STATE Update $\hat{p}_{k}^{(t)}$ using Eq.~(\textcolor{red}{\ref{Equation:Momentum-p}})\;
            
            \STATE Update $[k] \leftarrow (\mathcal{D}_{k}, \hat{p}_{k}^{(t)})$ \hfill{\textcolor{blue}{$\triangleright$  Repackaging}}
            
            \STATE Obtain the local filter $\mathcal{W}_{k}^{(t)}$ using Algorithm~\textcolor{red}{\ref{PseudoCode:Local-Model-Training}}\;
            
            \STATE Upload $\theta_{k}^{(t)}$ and $\mathcal{W}_{k}^{(t)}$ to the server\;
            
        }\ENDFOR
        
        \STATE Update the network ${\theta}^{(t+1)}$ using Eq.~(\textcolor{red}{\ref{Equation:FedAvg}})\;
        
        \FOR{$k \in \mathcal{S}$}{
        
            \STATE  $\mathcal{W}_{k} \leftarrow \mathcal{W}_{k}^{(t)}$\;
            
        }\ENDFOR
        
        \STATE Update the global filter ${\mathcal{W}}^{(t+1)}$ using Eq.~(\textcolor{red}{\ref{Equation:M-Step-at-Server}})\;
        
        }\ENDFOR
        
    \end{algorithmic}
    \caption{The training procedure of FedDiv}
    \label{PseudoCode:FedDiv}
\end{algorithm}
    
\begin{algorithm}[t]
\small
\textbf{Input}: {$\mathcal{D}_{k}$; $\theta^{(t)}_{k}$; $\mathcal{W}^{(t)}$}\\
\textbf{Output}: {Optimal local filter parameters $\mathcal{W}_{k}^{(t)}$}

\begin{algorithmic}[1]

    \STATE Initialize $\mathcal{W}_{k}^{(t)}=(\boldsymbol{\mu}_{k}^{(t)}, \boldsymbol{\sigma}_{k}^{(t)}, \boldsymbol{\pi}_{k}^{(t)})$ using $\mathcal{W}^{(t)}$
    
    \WHILE{$\mathcal{W}_{k}^{(t)}$ \textit{is not converged}}{
    
        \item[]  \textcolor{blue}{// E step:}
        
        \STATE ${\gamma}_{kg}(x,y;{\theta}_{k}^{(t)}) =  \frac{\pi_{kg}^{(t)}\cdot\mathcal{N}(\ell(x,y;\theta^{(t)}_{k});{{{\mu}}^{(t)}_{kg}},{{{\sigma}}^{(t)}_{kg}})}{\sum_{g^{\prime}=1}^{2}  \pi_{kg^{\prime}}^{(t)}\cdot\mathcal{N}(\ell(x,y;\theta^{(t)}_{k});{{{\mu}}^{(t)}_{kg^{\prime}}},{{{\sigma}}^{(t)}_{kg^{\prime}}})}$
        
        \item[]  \textcolor{blue}{// M step:}
        
        \STATE ${\mu}_{kg}^{(t)}=\frac{\sum_{(x, y) \in \mathcal{D}_k} {\gamma}_{kg}(x,y;{\theta}_{k}^{(t)}) \cdot \ell(x,y;\theta^{(t)}_{k})}{\sum_{{(x, y)} \in \mathcal{D}_k} {\gamma}_{kg}(x,y;{\theta}_{k}^{(t)})} $
        
        \STATE ${\sigma}_{kg}^{(t)}=\frac{\sum_{(x, y) \in \mathcal{D}_k} {\gamma}_{kg}(x,y;{\theta}_{k}^{(t)}) \cdot {[\ell(x,y;\theta^{(t)}_{k})-{\mu}_{kg}^{(t)}]}^{2}}{\sum_{{(x, y)} \in \mathcal{D}_k} {\gamma}_{kg}(x,y;{\theta}_{k}^{(t)})} $
        
        \STATE ${\pi}_{kg}^{(t)}=\frac{\sum_{{(x, y)} \in \mathcal{D}_k} {\gamma}_{kg}(x,y;{\theta}_{k}^{(t)})}{n_k}$
        
    }\ENDWHILE
    
\end{algorithmic}
\caption{Local filter training}
\label{PseudoCode:Local-Model-Training}
\end{algorithm}

\section{Additional Experimental Setups}

\noindent
\textbf{Non-IID data partitions.}
We employ Dirichlet distribution~\cite{lin2020ensemble}  with the fixed probability $p$ and the concentration parameter $\alpha_{Dir}$ to construct non-IID data partitions. Specifically, we begin by introducing an indicator matrix $\Phi \in {\mathbb{R}}^{C \times K}$, and each entry ${\Phi}_{ck}$ determines whether the $k$-th client has samples from the $c$-th class. For every entry, we assign a 1 or 0 sampled from the Bernoulli distribution with a fixed probability $p$. For the row of the matrix $\Phi$ that corresponds to the $c$-th class, we sample a probability vector $q_c\in {\mathbb{R}}^{{Q}_{c}}$ from the Dirichlet distribution with a concentration parameter $\alpha_{Dir} > 0$, where ${Q}_{c}=\sum_{k} {\Phi}_{ck}$. Then, we assign the ${k}^{\prime}$-th client a $q_{c{k}^{\prime}}$ proportion of the samples that belong to the $c$-th category, where ${k}^{\prime}$ denotes the client index with ${\Phi}_{ck}=1, k=1, \cdots, K$, and $\sum_{k^{\prime}=1}^{|q_c|} q_{c{k}^{\prime}}=1$.

\noindent
{\bf Additional Implementation Details.} 
Similar to \texttt{FedCorr}~\cite{xu2022fedcorr}, we select ResNet-18~\cite{resnet}, ResNet-34~\cite{resnet} and~Pre-trained ResNet-50~\cite{resnet} as the network backbones for CIFAR-10, CIFAR-100 and Clothing1M, respectively. During the local model training sessions, we train each local client model over $T=5$ local training epochs per communication round, using an SGD optimizer with a momentum of 0.5 and a mini-batch size of 10, 10, and 16 for CIFAR-10, CIFAR-100, and Clothing1M, respectively. 
For each optimizer, we set the learning rate as 0.03, 0.01, and 0.001 on CIFAR-10, CIFAR-100, and Clothing1M, respectively. In addition, during data pre-processing, the training samples are first normalized and then augmented by hiring random horizontal flipping and random cropping with padding of 4. For most thresholds conducted on the experiments, we set them to default as in \texttt{FedCorr}~\cite{xu2022fedcorr}, e.g.  {$\hat{\delta}_{k}=0.1$} in {Eq. (\ref{Equation:Data-Split})}, the probability of a sample being clean/noisy {$= 0.50$} in {Eq. (\ref{Equation:Label-Splitting})}, {$\xi=0.5$} in {Eq. (\ref{Equation:Logit-Debias})}, etc. Additionally, we determine {$\zeta$} in{ Eq. (\ref{Equation:Relabel})} using a small validation set, where $\zeta=0.70$ meets the peak of validation accuracies.

\noindent
\textbf{How to set $\omega$, $\mathcal{T}$ and $\mathcal{T}_{wu}$.}
In this work, we streamlined the multi-stage F-LNL process proposed in \texttt{FedCorr}~\cite{xu2022fedcorr} into a one-stage process, avoiding the complexity of executing multiple intricate steps across different stages as in \texttt{FedCorr}. However, for fair comparisons, we maintained an equivalent number of communication rounds as in \texttt{FedCorr}. This totals $\mathcal{T}_{wu}$, encompassing federated pre-processing from \texttt{FedCorr}'s training iterations, and $\mathcal{T}$, covering both federated fine-tuning and usual training stages involving \texttt{FedCorr}. Notably, within $\mathcal{T}_{wu}$, we solely utilize MixUp regularization~\cite{zhang2018mixup} to warm up the local neural network models for faster convergence.

Below, we will begin by introducing the multi-stage F-LNL pipeline proposed by \texttt{FedCorr}, followed by an analysis of fraction scheduling and the construction of the training rounds of \texttt{FedDiv}.
  
\begin{itemize}
    \item  \textbf{\texttt{FedCorr}.}  \texttt{FedCorr} comprises three FL stages: federated pre-processing, federated fine-tuning, and federated usual training. During the pre-processing stage, the FL model is initially pre-trained on all clients for $\mathcal{T}_{wu}$ \textit{iterations} (not training round) to guarantee initial convergence of model training. At the same time, \texttt{FedCorr} evaluates the quality of each client's dataset and identifies and relabels noisy samples. After this stage, a dimensionality-based filter~\cite{LID} is proposed to classify clients into clean and noisy ones. In the federated fine-tuning stage, \texttt{FedCorr} only fine-tunes the global model on relatively clean clients for $\mathcal{T}_{ft}$ rounds. At the end of this stage, \texttt{FedCorr} re-evaluates and relabels the remaining noisy clients. Finally, in the federated usual training stage, the global model is trained over $\mathcal{T}_{ut}$ rounds using \texttt{FedAvg}~\cite{FedAvg} on all the clients, incorporating the labels corrected in the previous two training stages.

    \item  \textbf{Fraction scheduling and communication rounds of \texttt{FedDiv}.} During FL training, a fixed fraction of clients will be selected at random to participate in local model training at the beginning of each round. Here, we set a fraction parameter $\omega$ to control the fraction scheduling, which is the same as the fine-tuning and usual training stages in \texttt{FedCorr}. However, during the pre-processing stage of \texttt{FedCorr}, every client must participate in local training exactly once in each iteration. Hence, any one client may be randomly sampled from all the clients with a probability of $\frac{1}{K}\cdot100\%$ to participate in local model training, without replacement. As three stages of \texttt{FedCorr} have been merged into one in \texttt{FedDiv}, to ensure fairness in training, we convert the training iterations of the pre-processing stage into the training rounds we used, which gives us the corresponding training rounds $\mathcal{T}_{wu}^{\prime}=(\mathcal{T}_{wu}\times K)/({w}\times K)=\mathcal{T}_{wu}/{w}$. Therefore, in our work, the total number of communication rounds in the entire training process is $\mathcal{T}_{wu}^{\prime}+\mathcal{T}$, where $\mathcal{T}=\mathcal{T}_{ft}+\mathcal{T}_{ut}$.
\end{itemize}

\begin{figure}[t]
    \centering

    \includegraphics[width=0.475\textwidth]{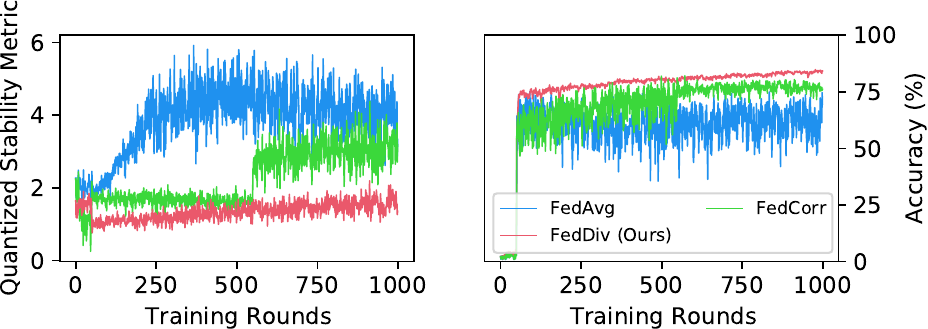}

    \caption{The evolution of quantized training stability \textit{v.s.} test classification performance across epochs for various F-LNL algorithms. We quantitatively assess training stability in F-LNL by computing the average proximal regularization metric~\cite{FedProx, xu2022fedcorr} between the weights of local and global neural network models in the current training round. The experiments are conducted on CIFAR-10 with $(p, \alpha_{Dir})=(0.3, 10.0)$ and $(\rho, \tau)=(6.0, 0.5)$.
    } 
    \label{Figure:Instability}
\end{figure}

\section{Additional Analysis}

\noindent
\textbf{Quantized training stability.}
To better grasp the motivation behind this approach, we propose using ``Quantized training stability'' to quantify the impact of data heterogeneity and noise heterogeneity~\cite{kim2022fedrn,RoFL} on the training instability experienced during local training sessions. Technically, quantized training stability can be measured by the average proximal regularization metric between local and global model weights, denoted as $\theta_{k}^{(t)}$ and $\theta^{(t)}$ respectively, at round $t$. This is calculated by $\frac{1}{|\mathcal{S}|}\sum_{k\in\mathcal{S}} |\theta_{k}^{(t)}- \theta^{(t)}|^{2}$. As depicted in Figure~\textcolor{red}{\ref{Figure:Instability}}, this instability results in notable discrepancies in weight divergence between local and global models, potentially hindering the performance enhancement of the aggregated model if left unaddressed. Additionally, considering the efficacy of different noise filtering strategies, our proposed federated noise filtering demonstrates superior performance in label noise identification per client, leading to decreased training instability during local training sessions and thus achieving higher classification performance of the aggregated model.

\begin{figure*}[t]
    \centering
    \begin{subfigure}{\textwidth}
        \centering
        \includegraphics[width=0.75\textwidth]{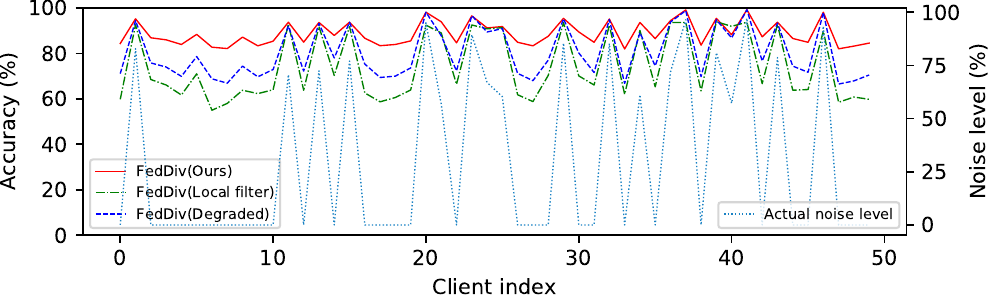}
    \end{subfigure}
    \begin{subfigure}{\textwidth}
        \centering
        \includegraphics[width=0.75\textwidth]{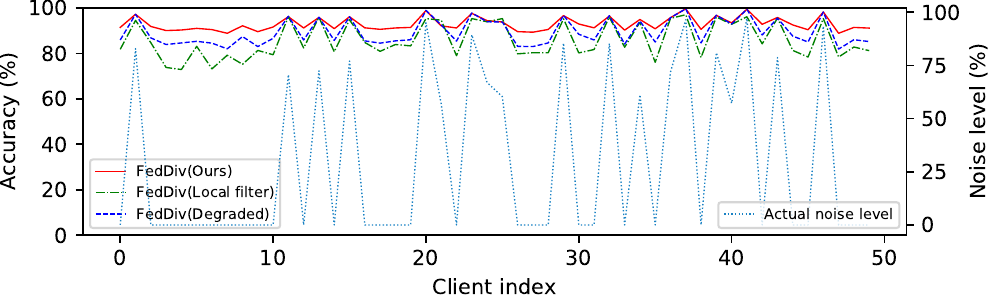}
    \end{subfigure}

    \caption{The accuracy of noisy label identification vs. different clients. 
    The lightblue dotted line represents the actual noise level of each client, while the (dotted) lines in deep bright colors indicate the noise filtering performance with respect to different noise filters. The experiment is conducted on CIFAR-100 with the IID data partition under the noise setting $(\rho, \tau)=(0.4, 0.5)$. TOP: Evaluation in the $50$-th communication round of the usual training stage; BOTTOM: Evaluation in the $500$-th communication round of the usual training stage.}
    \label{Figure:Filtering-Accuracy}
\end{figure*}

\begin{figure*}[t]
    \centering
    \begin{subfigure}{\textwidth}
        \centering
        \includegraphics[width=0.75\textwidth]{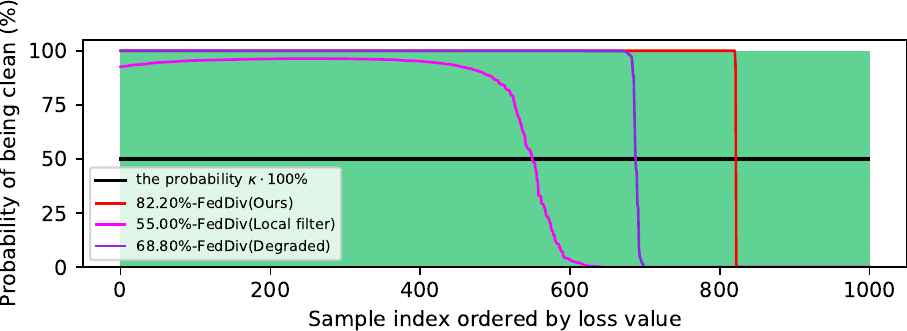}
    \end{subfigure}
    \begin{subfigure}{\textwidth}
        \centering
        \includegraphics[width=0.75\textwidth]{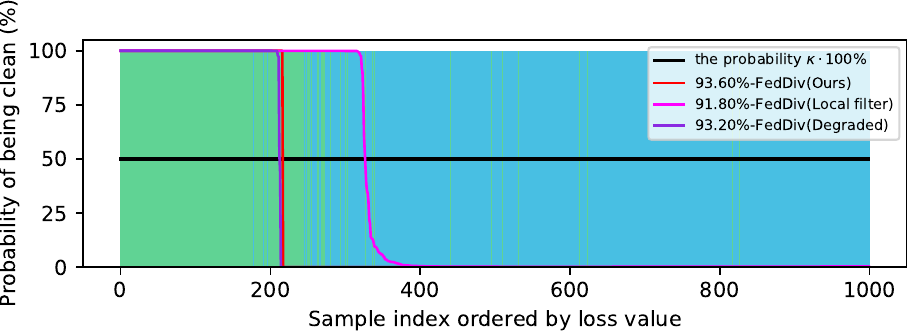}
    \end{subfigure}

    \caption{Two examples to illustrate the performance of three noise filters to identify label noise on a clean client (with the lowest accuracy of label noise filtering) and a noisy client. We show the probability distributions of samples being clean predicted by each filter, with the samples ranked according to the per-sample loss function values. In the legend, the percentages show the accuracy of noisy label identification with respect to each filter. In addition, as illustrated by the black line, a sample that is considered clean should have a predicted probability higher than $\kappa\cdot100\%$. Furthermore, the green and blue bars represent, respectively, the distribution of the given \textbf{\textcolor{green}{clean}} and \textbf{\textcolor{blue}{noisy}} samples. The evaluation is performed at the end of the 50-th communication round in the usual federated training stage. The experiment is conducted on CIFAR-100 with the IID data partition under the noise setting $(\rho, \tau)=(0.4, 0.5)$. TOP: Evaluation on the clean client; BOTTOM: Evaluation on the noisy client with $\delta=0.725$. (\textbf{Best viewed zoomed in.})}
    \label{Figure:Performance-On-Different-Filter}
\end{figure*}

\noindent
\textbf{Further evaluation of federated noise filtering.} To further verify the capability of our proposed label noise filtering strategy, we again compare \texttt{FedDiv} with \texttt{FedDiv(Local\;filter)} and \texttt{FedDiv(Degraded)} in Figure~\textcolor{red}{\ref{Figure:Filtering-Accuracy}} and Figure~\textcolor{red}{\ref{Figure:Performance-On-Different-Filter}}. Both experiments are conducted on CIFAR-100 with the noise setting $(\rho, \tau)=(0.4, 0.5)$ for the IID data partition. Specifically, Figure~\textcolor{red}{\ref{Figure:Filtering-Accuracy}} shows the accuracy of label noise filtering over all 50 clients at different communication rounds, while Figure~\textcolor{red}{\ref{Figure:Performance-On-Different-Filter}} provides two examples to illustrate the noise filtering performance of different noise filters on clean and noisy clients.

As depicted in both Figure~\textcolor{red}{\ref{Figure:Filtering-Accuracy}} and Figure~\textcolor{red}{\ref{Figure:Performance-On-Different-Filter}}, the proposed strategy consistently produces stronger label noise filtering capabilities on the vast majority of clients than the alternative solutions. Additionally, Figure~\textcolor{red}{\ref{Figure:Filtering-Accuracy}} also shows that all these three noise filtering schemes significantly improve the label noise identification performance as model training proceeds, especially on clean clients—possibly because the network model offers greater classification performance—but ours continues to perform the best.
These results once again highlight the feasibility of our proposed noise filtering strategy.

\begin{figure*}[t]
  \centering

    \begin{tabular}{cccccccc}
    \multicolumn{2}{c}{\multirow{10}[0]{*}{Given}} & \multicolumn{6}{c}{\multirow{15}[0]{*}{\includegraphics[width=0.875\textwidth]{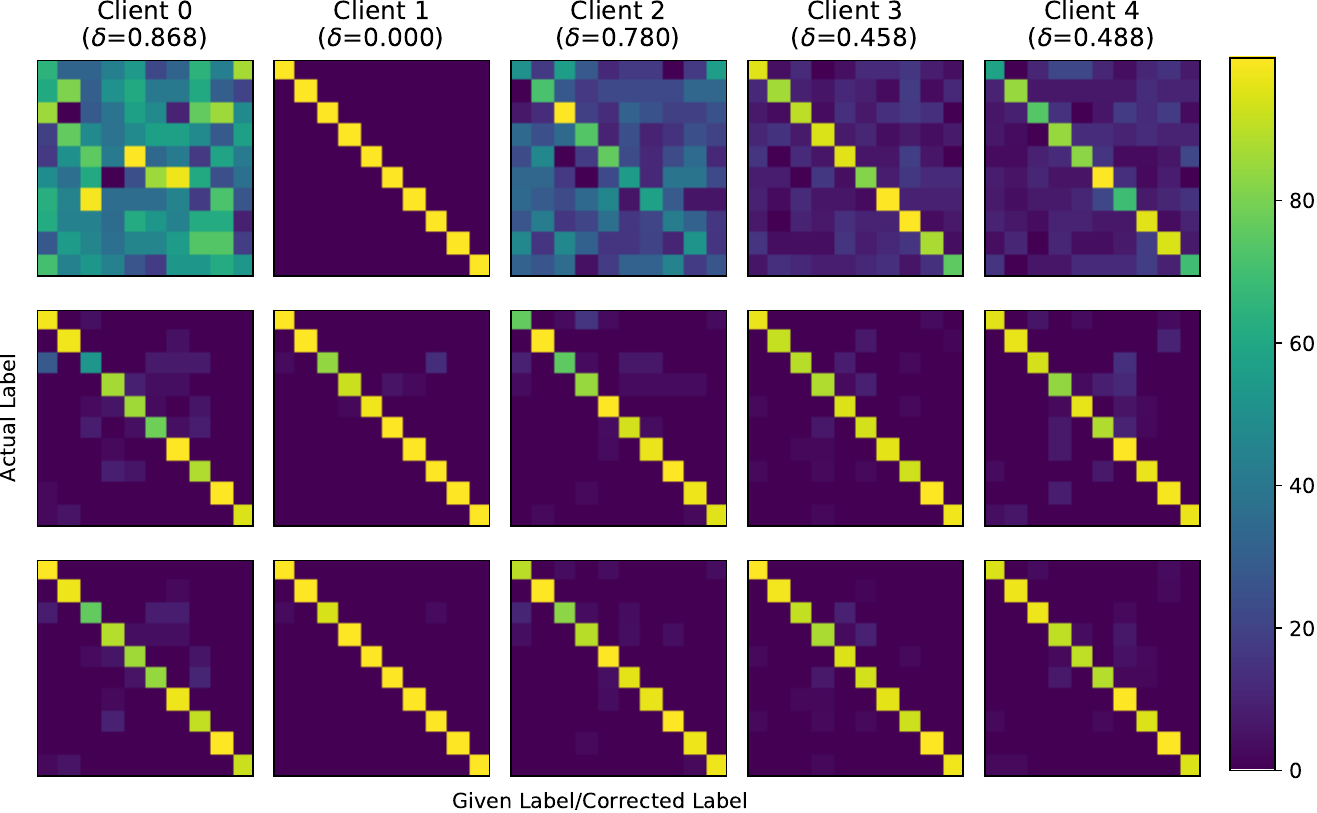}}} \\
    \multicolumn{2}{c}{} & \multicolumn{6}{c}{} \\
    \multicolumn{2}{c}{} & \multicolumn{6}{c}{} \\
    \multicolumn{2}{c}{} & \multicolumn{6}{c}{} \\
    \multicolumn{2}{c}{} & \multicolumn{6}{c}{} \\
    \multicolumn{2}{c}{} & \multicolumn{6}{c}{} \\
    \multicolumn{2}{c}{\multirow{13}[0]{*}{Thread 1}} & \multicolumn{6}{c}{} \\
    \multicolumn{2}{c}{} & \multicolumn{6}{c}{} \\
    \multicolumn{2}{c}{} & \multicolumn{6}{c}{} \\
    \multicolumn{2}{c}{} & \multicolumn{6}{c}{} \\
    \multicolumn{2}{c}{} & \multicolumn{6}{c}{} \\
    \multicolumn{2}{c}{} & \multicolumn{6}{c}{} \\
    \multicolumn{2}{c}{\multirow{15}[0]{*}{Thread 2}} & \multicolumn{6}{c}{} \\
    \multicolumn{2}{c}{} & \multicolumn{6}{c}{} \\
    \multicolumn{2}{c}{} & \multicolumn{6}{c}{} \\
    \multicolumn{2}{c}{} & \multicolumn{6}{c}{} \\
    \multicolumn{2}{c}{} & \multicolumn{6}{c}{} \\
    \multicolumn{2}{c}{} & \multicolumn{6}{c}{} \\
    \multicolumn{2}{c}{} & \multicolumn{6}{c}{} \\
    \multicolumn{2}{c}{} & \multicolumn{6}{c}{} \\
    \multicolumn{2}{c}{} & \multicolumn{6}{c}{} \\
    \multicolumn{2}{c}{} & \multicolumn{6}{c}{} \\
    \multicolumn{2}{c}{} & \multicolumn{6}{c}{} \\
    \multicolumn{2}{c}{} & \multicolumn{6}{c}{} \\
    \multicolumn{2}{c}{} & \multicolumn{6}{c}{} \\
    \multicolumn{2}{c}{} & \multicolumn{6}{c}{} \\
    \end{tabular}%

    \caption{An evaluation of label noise filtering, noisy sample relabeling and labeled sample re-selection on five representative clients in the proposed \texttt{FedDiv}.  As indicated by the heat maps, three confusion matrices for each client are associated to the actual labels v.s. the given labels before processing, the corrected labels after label noise filtering and noisy sample relabeling (named Thread 1), and the corrected labels after labeled sample re-selection (named Thread 2), respectively.
    Note that, in practice, noisy label relabeling and labeled sample re-selection may not necessarily be conducted on clean clients (e.g., Client 1) during local model training, in accordance with Eq. (\textcolor{red}{10}).
    The experiment is conducted on CIFAR-10 with the IID data partition under the noise setting $(\rho, \tau)=(0.8, 0.5)$. }

  \label{Figure:Heatmap}%
\end{figure*}

\noindent
\textbf{Further evaluation of filtering, relabeling and re-selection.} To emphasize each \texttt{FedDiv} thread's effectiveness in label noise filtering, noisy sample relabeling, and labeled sample re-selection, we compare confusion matrices before processing, after label noise filtering and noisy sample relabeling (Thread 1), and after labeled sample re-selection (Thread 2) in Figure~\textcolor{red}{\ref{Figure:Heatmap}}. Figure~\textcolor{red}{\ref{Figure:Heatmap}} displays heat maps of these three confusion matrices on five representative clients. On clean or noisy clients with varying noise levels, each thread gradually eliminates label noise, confirming the performance of each \texttt{FedDiv} component.

\begin{figure*}[t]
    \centering
    \begin{subfigure}{.45\textwidth}
        \centering
        \includegraphics[width=0.75\textwidth]{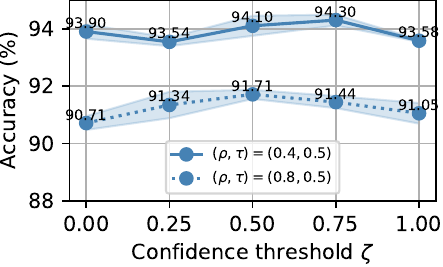}
        \caption*{(a) IID}
    \end{subfigure}
    \begin{subfigure}{.45\textwidth}
        \centering
        \includegraphics[width=0.75\textwidth]{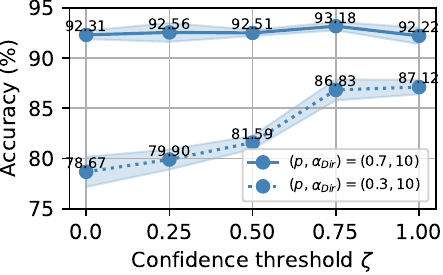}
        \caption*{(b) Non-IID}
    \end{subfigure}

    \caption{Sensitivity with respect to the hyper-parameter $\zeta$. We show the test accuracy to illustrate the classification performance of the final FL model when we set $\zeta$ to $0.00$, $0.25$, $0.50$, $0.75$, and $1.00$, respectively. We conduct these experiments on CIFAR-10 with both IID and non-IID data partitions. For the non-IID data partition, the noise level is set to $(\rho, \tau)=(0.6, 0.5)$.}
    \label{Figure:Hyper-Parameter-Sensitivity}
\end{figure*}

\noindent
\textbf{Hyper-parameter sensitivity.} 
We analyze hyper-parameter sensitivity to the confidence threshold $\zeta$. As shown in Figure~\textcolor{red}{\ref{Figure:Hyper-Parameter-Sensitivity}}, the proposed approach consistently achieves higher classification performance when $\zeta$ is set to 0.75. Therefore, $\zeta=0.75$ is an excellent choice for setting the confidence threshold for noisy sample relabeling when training the FL model on CIFAR-10 and CIFAR-100 under different label noise settings for both IID and non-IID data partitions.

\end{document}